%% file: topic_tracker_main.tex
\documentclass[preprint,review,  a4paper]{elsarticle}

\usepackage{lineno,hyperref}
\modulolinenumbers[5]
\usepackage{url}
\usepackage{amsmath}
\usepackage{amssymb}
\usepackage{amsthm}
\usepackage{mathrsfs}
\usepackage{graphicx}
\usepackage{todonotes}
\usepackage{multirow}
\usepackage{array}
\usepackage[linesnumbered,ruled,vlined]{algorithm2e}
\usepackage{chngpage}
\usepackage{booktabs}
\usepackage{xcolor}
\usepackage[T1]{fontenc}
\usepackage[english]{babel}
\usepackage{tabu}
\usepackage{subcaption}
\usepackage{longtable}
\usepackage{xspace}
\usepackage{microtype}

\usepackage{enumitem}
\usepackage{adjustbox}
\usepackage{wasysym}
\usepackage[utf8]{inputenc}
\usepackage{geometry}
\usepackage{blkarray}
\usepackage{float}
\restylefloat{table}
\usepackage{pifont}%
\usepackage{makecell}
\usepackage{hyperref}

\journal{SoftwareX}

\newcommand{\ttv}{{\fontfamily{lmss}\selectfont TopicTracker}\xspace}

\newcommand{\mindead}{\textsf {min\_dead}\xspace}
\newcommand{\minreborn}{\textsf {min\_reborn}\xspace}
\newcommand{\mintes}{\textsf {min\_tes}\xspace}

\newcommand{\mc}{\mathcal}
\newcommand\var{\texttt}

\newcommand{\mb}{\mathbf}

\begin{document}

\begin{frontmatter}

\title{TopicTracker: A Platform for Topic Trajectory Identification and Visualisation}
% \tnotetext[mytitlenote]{Fully documented templates are available in the elsarticle package on \href{http://www.ctan.org/tex-archive/macros/latex/contrib/elsarticle}{CTAN}.}

\author[cor]{Yong-Bin Kang\corref{ybk}}
\ead{ykang@swin.edu.au}
% \author[cti]{Beth Webster}
% \ead{emwebster@swin.edu.au}
% \author[cti]{Alfons Palangkaraya}
% \ead{apalangkaraya@swin.edu.au}
% \author[csse]{Kai Qin}
% \ead{kqin@swin.edu.au}
\author[fb]{Timos Sellis\corref{ts}}
\ead{tsellis@fb.com}

\cortext[ybk]{Corresponding Author}
\cortext[ts]{Work done while at Swinburne University of Technology}
\address[cor] {Department of Media and Communication, Swinburne University of Technology, Australia}
% \address[cti] {Swinburne Business School, Swinburne University of Technology, Australia}
\address[fb] {Facebook, USA}

\begin{abstract}
Topic trajectory information provides crucial insight into the dynamics of topics and their evolutionary relationships over a given time. Also, this information can help to improve our understanding on how new topics have emerged or formed through a sequential or interrelated events of emergence, modification and integration of prior topics. 
Nevertheless, the implementation of the existing methods for topic trajectory identification is rarely available as usable software. In this paper, we present \ttv, a platform for topic trajectory identification and visualisation. The key of \ttv is that it can represent the three facets of information together, given two kinds of input: a time-stamped topic profile consisting of the set of the underlying topics over time, and the evolution strength matrix among them: evolutionary pathways of dynamic topics, evolution states of the topics, and topic importance.
%\ttv would also help the user to speed up the development of different models for topic trajectory identification more easily by tailoring to to specific related areas of interest.
\ttv is a publicly available software implemented using the R software.

\end{abstract}

\begin{keyword}
TopicTracker \sep topic trajectory \sep topic evolution  \sep topic tracking
\end{keyword}

\end{frontmatter}

\section*{Code metadata}
\label{}

\begin{table}[H]
\begin{tabular}{|l|p{6.5cm}|p{6.5cm}|}
\hline
C1 & Current code version & v1.0.0-beta \\
\hline
C2 & Permanent link to code/repository used for this code version &  \url{https://github.com/Yongbinkang/topicTracker}
% \yb{The permanent link to code/repository or the zip archive should include the following requirements: README.txt and LICENSE.txt. Source code in a src/ directory, not the root of the repository. Tag corresponding with the version of the software that is reviewed.Documentation in the repository in a docs/ directory, and/or READMEs, as appropriate.}
\\
\hline
C3 & Code Ocean compute capsule & \\
\hline
C4 & Legal Code License   & MIT License\\
\hline
C5 & Code versioning system used & git \\
\hline
C6 & Software code languages, tools, and services used & R \\
\hline
C7 & Compilation requirements, operating environments \& dependencies & R version $\geq$ 3.6, and R packages (igraph, hash, plotrix)\\
\hline
C8 & If available Link to developer documentation/manual &  \url{https://github.com/Yongbinkang/topicTracker} \\
\hline
C9 & Support email for questions & \makecell{\href{mailto:ykang@swin.edu.au}{\texttt{ykang@swin.edu.au}}, \\ \href{mailto:yongbin.kang@gmail.com}{\texttt{yongbin.kang@gmail.com}}}\\
\hline
\end{tabular}
\caption{Code metadata}
\label{tbl:code_metadata} 
\end{table}

% \linenumbers


\input{significance.tex}
\input{software_desc}
\input{illustration}

\input{impact}
\input{conclusion}
\input{ack}

\bibliography{ybk}

\end{document}

%% file: significance.tex
\section{Motivation and significance}
\label{sec:intro}

Topic trajectory identification is a research area that has attracted significant attention from scientific institutions and innovation-industry sectors. 
In this area, a fundamental is the use of \textit{topic modeling} to discover latent thematic topics (or concepts) from a document collection, where each topic consists of its representative terms extracted from the collection~\cite{Lee:2017}.
More recently, dynamic topic modeling has also been utilised to identify topics and their evolution over a time period~\cite{Zhang:2015, Greene:2016, Song:2016}. 
Identifying topic trajectories provides precious insights into the dynamics of topics over time. For example, in scientific and patented innovation domains, such trajectories can significantly help to
distinguish outstanding research or technological topics, and discover their evolutionary pathways reflecting how new topics have been emerged or formed through a sequential of the events of the emergence, modification and integration of past topics~\cite{Zhou:2017, Song2014, Yoon:2011, Zhang:2017}. We view a \textit{trajectory of topics} as a main stream or an evolutionary pathway of topics over time\footnote{To simplify the presentation, we do not distinguish between `trajectory' and 'evolutionary pathways' and use them interchangeably.}. Also, we define an \textit{evolutionary pathway} as a series of evolutionary relationships between older topics and newer topics. 
% For political agenda analysis, given a series of political speech records in text, identifying topic trajectory can uncover what political agenda topics have been evolved over time, and what agenda topics have driven to give rise to new political agenda topics over time~\cite{Greene:2016}. In an e-commerce domain, topic trajectory can track a temporal evolution of market-competitive product-related topics (e.g. luxury goods) over time in a brand sale market~\cite{Zhang:2015}.
%Also, in life-science research, identifying topic trajectory of transmission events (that can be seen as topics) of an outbreak of an infectious disease can provide significant insights into inferring who infected whom~\cite{Jombart:2018}. 
Consequently, due to this merit, many studies were conducted in recent years for designing different methods for topic trajectory identification (see  Table~\ref{tbl:related_work}).

% In topic trajectory identification, a key objective is to identify credible \textit{trajectories} of \textit{non-contemporary topics} (simply topics) over a given fixed time period (simply time). We define a trajectory of topics as a main stream or evolutionary pathway of topics over time~\footnote{To simplify the presentation, we do not distinguish between phrases `trajectory' and 'evolution pathways' and use them interchangeably.}. Also, we define an evolutionary pathway as a series of an evolutionary relationship between an old topic and a new topic.

\newcolumntype{P}[1]{>{\centering\arraybackslash}p{#1}}
\newcommand{\cmark}{\ding{51}}%
\newcommand{\xmark}{\ding{55}}%

\begin{table*}[t!]\small
\begin{center}
\caption{Previous studies on topic trajectory identification and software availability.}\label{tbl:related_work}
\scalebox{0.9} {
\setlength{\extrarowheight}{1.5pt}
\setlength{\tabcolsep}{2pt}
\begin{tabular}{lcl}
\hline
{\bf Study} & 
{\bf Software availability} & 
{\bf Domain}\\\hline
He et al. (2009)~\cite{He:2009} 
& \xmark & \multirow{6}{*}{Scientific literature (topic: technology/knowledge concept)}
\\ 
Jo et al. (2011)~\cite{Jo:2011}
& \xmark & \\
Song et al. (2014)~\cite{Song2014}
& \xmark  &\\ 
Zhou et al. (2017)~\cite{Zhou:2017}
& \xmark  & \\ 
 Zhang et al. (2017)~\cite{Zhang:2017}
& \xmark  & \\ 
Jung et al. (2020)~\cite{Jung:2020}
& \xmark & \\  \hline
Yoon et al. (2011)~\cite{Yoon:2011} & \xmark & \multirow{7}{*}{Patents (topic: technology/knowledge concept)} \\ 
Zhong  et al. (2016)~\cite{Zhong:2016} & \xmark &\\ 
 Lee et al. (2017)~\cite{Lee:2017} & \xmark&\\  
 Park et al. (2017)~\cite{Park:2017} & \xmark &\\  
 Triulzi et al. (2020)~\cite{Triulzi:2020} & \xmark &\\  
 Huang et al. (2020)~\cite{Huang:2020} & \xmark &\\  
 Qiu et al. (2020)~\cite{Qiu:2020} & \xmark &\\  
\hline
Zhang  et al. (2015)~\cite{Zhang:2015} & \xmark &{E-commerce (topic: market brand)} \\  \hline
Greene  et al. (2016)~\cite{Greene:2016} & \xmark& {Politics (topic: political agenda)}\\  \hline
Song et al. (2016)~\cite{Song:2016} & \xmark & History (topic: historical event)\\  \hline
Gaul et al. (2017)~\cite{Gaul:2017} & \xmark & Online news (topic: online news) \\
\hline
\end{tabular}}
\end{center}
\end{table*}

Unfortunately, the implementation of these methods 
to encourage their use by the wider scientific community that are interested in topic trajectory identification has been still remained limited. 
Primarily, a large body of the methods is not available as readily usable software as seen in Table~\ref{tbl:related_work}. 
In topic modeling and its application areas, existing software tools were developed with little consideration for preparing and formatting the data for topic trajectory identification in a simple and easy way to use. 
This requires the users to directly implement their algorithms for topic trajectory identification from the results of topic modeling. Further, this leads to unnecessary time spent for data preparation and limits effective comparison of results produced by different topic trajectory identification models as well.

To address these issues, we have developed \ttv,  a platform for topic trajectory identification and visualisation.
%\ttv would also help the user to speed up the development of different models for topic trajectory identification more easily by tailoring to to specific related areas of interest.
\ttv is a software implemented using the R software.
% In the following sections, we present the details of \ttv's design, its illustrative example, and its contributions to the community for topic trajectory identification.

%% file: software_desc.tex
\section{Software description}
\label{sec:software}
We present the architecture of \ttv and its main functionalities.

\subsection{Software Architecture}
\label{sec:soft_arch}

\begin{figure*}[!h]
    \includegraphics[trim=0cm 0cm 0cm 0cm, clip, width=380pt]{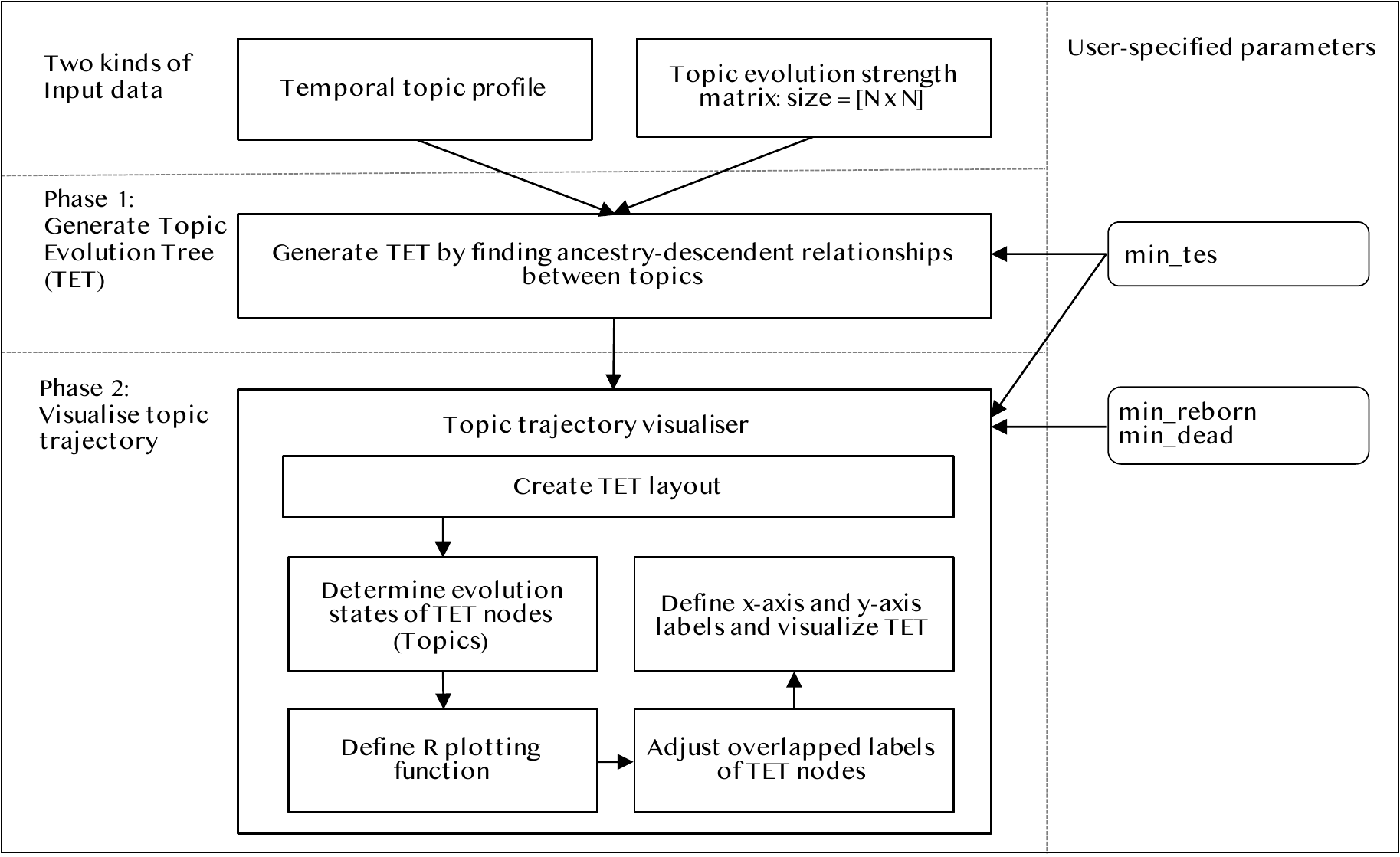}
    \caption{The architecture of \ttv.
    % The architecture of \ttv: (1) N: the number of topics; (2) \minevol: the minimum evolution strength between topics that will be used to find possible ancestors of topics. The evolution strengths $\leq$ \minevol will not be considered in finding the ancestors of topics; (3) \minreborn:  the minimum time period of topics which has been elapsed between the moment of emergence of topics until their evolved topics appear; (3) \mindead: the minimum time period of topics that goes into unobserved; and (4) \mingnore: the minimum evolution strength that will be ignored in visualising TET.
    }
    \label{fig:ttv_overview}
\end{figure*}

The architecture of \ttv's code design is depicted in Fig.~\ref{fig:ttv_overview}. 
It is designed to distinguish the inferential module for building a \textit{Topic Evolution Tree} (TET) of topics (Phase 1) from the code for visualising their topic trajectories (Phase 2). One key idea in designing \ttv is to discover evolutionary pathways (i.e., topic trajectories) between non-contemporary topics by constructing their most likely genealogy tree (i.e., TET) over a given time. Because of this flexibility, one can easily customise the code to modify or define a new shape of a TET.

The following summarises the workflow within the architecture: 
\begin{itemize}
    \item First, two kinds of input data must be provided: a \textit{temporal topic profile} comprising of the profile of the underlying time-stamped topics, and a $N \times N$ \textit{Topic Evolution Strength} (TES) matrix, where $N$ denotes the number of the topics in the profile. 
    
    \item Second, \ttv infers the TET of the topics in the profile with a user-specified parameter, \mintes (Phase 1), the minimum TES among the topics  to find their possible ancestors. The topics with TESs less than \mintes will not be considered in finding their ancestors. Given a topic $v$ in the TET, its pathway to the most ancestral topic indicates the trajectory of $v$. Thus, TET is a backbone reflecting evolutionary pathways of topics.
    \item Third, \ttv visualises the inferred TET with the five central modules (Phase 2) using  three parameters: \minreborn is the minimum time period of topics which has been elapsed between the moment of their emergence until their evolved topics appear; \mindead is the minimum time period of topics that goes into unobserved; and \mintes.
\end{itemize}

\begin{figure}[!h]
\centering
\includegraphics[trim=0cm 0cm 0cm 0cm, clip, width=300pt]{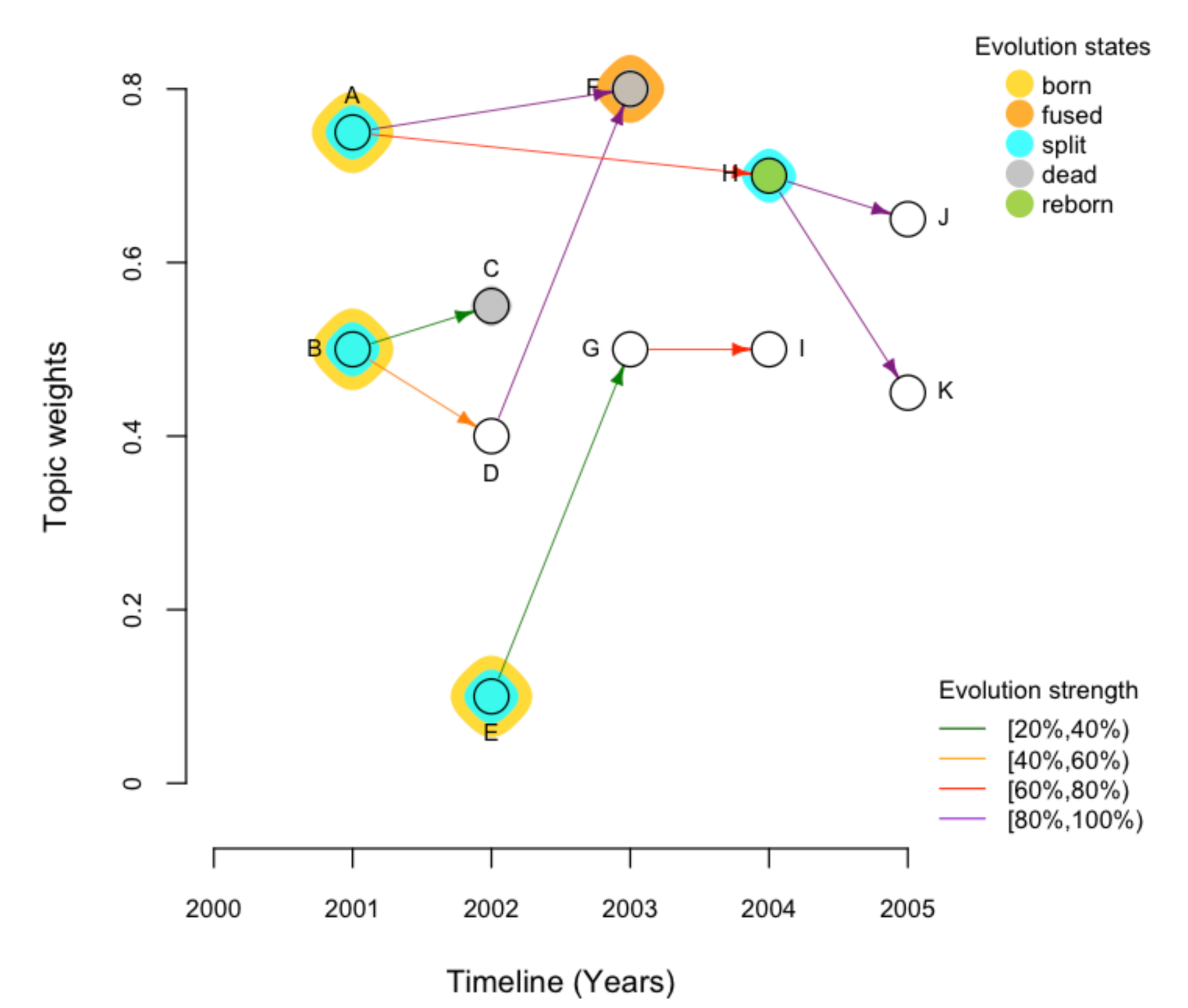}
\caption{An example TET generated by \ttv. 
% By default, uncolored topics denote {flourishing} topics. 
% 
The parameters are set as: \minreborn = 2 years, \mindead = 1 year, and \mintes = 0.2. The explanation about this example is provided in Section~\ref{sec:examples}.} \label{fig:tet_example}
\end{figure}

An example TET is presented in Fig.~\ref{fig:tet_example}, where each node denotes a topic. 
The evolutionary relationship between two topics is denoted by a directed edge, and its TES is represented by a different edge color. The evolution state of each node is marked with a different node color. The y-axis shows  the importance of topics normalised in [0,1]. We emphasize that the {integration of \textit{three information facets} in TET} provides insight about topic trajectories: the evolutionary pathways of topics with their TESs over time, evolution states of topics over time, and topic importance. Below we elaborate the construction process of a TET. 
% The more detailed illustration of TET will be discussed in Section~\ref{sec:traj_detail}.

\subsection{Software Functionalities}
\label{sec:soft_comp}
We now present the detailed descriptions about the input data and the two phases embodied in \ttv.

\subsubsection{Input data format}
The two kinds of input data\footnote{Discussion of generating this input data is beyond the scope of this paper.} must be given to run \ttv: a temporal topic profile $\mc{P}$, and a TES matrix $\mb{M}$.
$\mc{P}$ contains the descriptions about the target topics whose trajectories will be generated. An example of $\mc{P}$ is given in Table~\ref{tbl:P}, and it has the following fields for each topic $v$:
\begin{itemize}
    \item \texttt{id} is the unique identifier of $v$. 
    \item \texttt{index} is the unique integer index of $v$ (starting from 0).
    \item \texttt {weight} is the  weight of $v$ given each year.
    \item \texttt{year} (in format yyyy) is the year that $v$ was generated.
    % \item \texttt{words} (e.g., using top-5 words: \texttt{[`opinion', `computer', `lab', `user', `human']}) are the top-$N$ words representing $v$.
    \item \texttt{words} are the top-$N$ words representing $v$.
    The information about \texttt {weight}, \texttt{year}, and \texttt{words} can be generated by a topic model.\footnote{For example, a dynamic topic model~\cite{Blei:2006} can generate such information. Discussion about generating such information is beyond the scope of this paper.}
    
    % \item \texttt{word\_weights} (e.g., using top-5 word\_weights: \texttt{[0.43, 0.62, 0.51, 0.64, 0.8]}) are the weights of the top-$N$ words, where all weights are normalised in [0,1]. The higher a weight is, it is considered to be more important. 
    % The information about \texttt {weight}, \texttt{year}, \texttt{words} and \texttt{word\_weights} can be generated by a topic model.\footnote{For example, a dynamic topic model~\cite{Blei:2006} can generate such information. Discussion about how to generate such information is beyond the scope of this paper.}
\end{itemize}

The $\mb{M}$ matrix is a $N \times N$ matrix, where $N$ is the number of the topics presented in $\mc{P}$, where:
\begin{itemize}
    \item The $i$-th row and $j$-th column of $\mb{M}$ represents the TES of the $i$-th row topic (old) towards the $j$-th column topic (new).
    
    \item A TES only exists between a pair of two \textit{non-contemporary} topics as we only estimate the TES between topics on different time slots.
    Thus, we set TESs between \textit{contemporary topics} as 0. Also,
    by default, we set the diagonal entries to be 1.
    \item All the topics are sorted in ascending order according to their time slots. Thus, the first is the most ancient topic and the last the most recent topic.
    \item $\mb{M}$ is a non-symmetric matrix, where all entries below the main diagonal do not hold any values, as we are only interested in the calculation of the TES between two topics $x$ and $y$, if only if $time(x) < time(y)$, where $time$ is the function returning the time slot of a given topic.
    \item All entries in $\mb{M}$ are normalised in [0,1], where the higher the more important.
\end{itemize}

An example of $\mb{M}$ is given in Table~\ref{tbl:M}.
To estimate TESs in $\mb{M}$, the concept of similarity has been widely used in most related works ~\cite{He:2009, Zhou:2017, Jung:2020,Qiu:2020, Lee:2017,Greene:2016}. The fundamental is to formulate a similarity measure between $x$ and $y$ using the aggregated similarities between $x$'s top-$k$ descriptive terms and $y$'s top-$k$ descriptive terms (often $K$=10)~\cite{Derek:2015,Greene:2016}. 
%Note  that  \ttv  does not  rely  on  a  particular method  for  creating the matrix $\mb{M}$.

\subsubsection{Creating TET layout}\label{sec:tet_layout}

% In a TET, topics are represented as nodes and their evolutionary relationships are denoted as edges. 
We construct a TET from the given $\mc{P}$ and $\mb{M}$.
The goal of constructing a TET is to identify the most likely genealogy of topics over a given time period. Technically, this problem is formulated as finding an optimum branching in a directed tree, where each direct edge connects a topic and the topics evolved from it. 
% We define TET as follows: 
Let $\text{TET} = (V, E)$ be a directed, weighted tree, where $V=\{v_1, ..., v_n\}$ is the set of nodes that correspond to the $n$ topics in $\mc{P}$. These topics are time-stamped meaning that these are collected according to the $k$ different time slots, where $k$ is the carnality of the years observed in $\mc{P}$. $E \subseteq V \times V$ represents the set of directed edges that reflect evolutionary relationships in $V$, such that $(v_i, v_j) \in E$ if only if $time(v_i) < time(t_j)$, where the $time$ function specifies the time slot of a given topic. An edge $(v_i,v_j)$ is an ordered pair of two nodes $v_i$ and $v_j$, and is interpreted as there is a directed dependency from an ancestry topic $v_i$ to a descendent topic $v_j$. Each edge is associated with its TES, drawn from $\mb{M}$. A TES reflects how strong each ancestry-descendent relationship is between two topics. 

The TET algorithm is implemented in the function \texttt{buildTES()} in \texttt{topicTracker.R}, and has the following bases:
\begin{enumerate}
    \item We assume that there is the dummy \textit{root} node in a TET to merely make a tree. Thus, the most ancient topic for each topic are connected to the root node.
    
    \item Ancestry topics always precede their descendant topics in time. 
    
    \item A topic $v$ can have no ancestor, meaning that $v$ is newly emerged. On this occasion, $v$ is connected to the root node. 
    
    \item Each topic $v$ can have multiple ancestors as more than one topics in the past can be assembled together into the emergence of $v$ in a later time.
    
    \item In a TET, evolutionary transitivity relationships are logically inferred by navigating the edges between past and new topics. Suppose that there are two evolutionary relationships: an older topic \var{`wireless technology'} influences on the generation of a recent topic \var{`mobile devices'}; and a topic \var{`mobile devices'} influences on the generation of a more recent topic \var{`Android phone'}. Then, we infer the relationship that \var{`wireless technology'} can also influence on the generation of \var{`Android phone'}.
    
    \item A complexity of a TET is determined by the number of edges connected in the TET. This number is determined by \mintes. That is, we only include all edges whose TESs are equal to and greater than \mintes. Also, in a TET, cycles are not observed as ancestries cannot go back in time. 
    
    % \item If the topics form more than one connected trees, we build a TET by connecting the oldest topic of each connected tree to the root node to form a single connected tree.
    
    \item An evolutionary pathway of a topic $v$ is defined as 
    a sequence of connected topics from $v$ to the root node. Given the same evolutionary pathway, among all possible parents $\var{Par}(v)$ of $v$, some are more likely than others. A parent is the immediate ancestor of a topic. 
    This likelihood is estimated based on the TESs, from $\mb{M}$, between $v$ and $\var{Par}(v)$. We choose the parent with the highest TES with $v$. 
    To illustrate, in Fig.~\ref{fig:multi_anc} (a),  given a topic $\var{F}$, there are three possible evolutionary paths: (1) $\{(\var{B}, \var{F})\}$, (2) $\{(\var{B},\var{C}), (\var{C},\var{F})\}$, and (3) $\{(\var{B},\var{D}), (\var{D},\var{F})\}$. Given the paths (1) and (2), there are two possible parents for $\var{F}$, $\var{B}$ and $\var{C}$, as these exist on the same pathway. In  this case, we choose $\var{C}$ as its TES with $\var{F}$ is higher than that of $\var{B}$. Thus, we do not connect $\var{F}$ to $\var{B}$. The TESs are indicated by different edge colors as presented in the legend. In the same manner, given the paths (1) and (3), we choose $\var{D}$ not $\var{B}$ as the parent of $\var{F}$. The result is given in Fig.~\ref{fig:multi_anc} (b).

    \begin{figure}[!h]
        \begin{minipage}[t]{.45\textwidth}
            \centering
            \includegraphics[trim=0cm 0cm 0cm 0cm, clip, width=200pt]{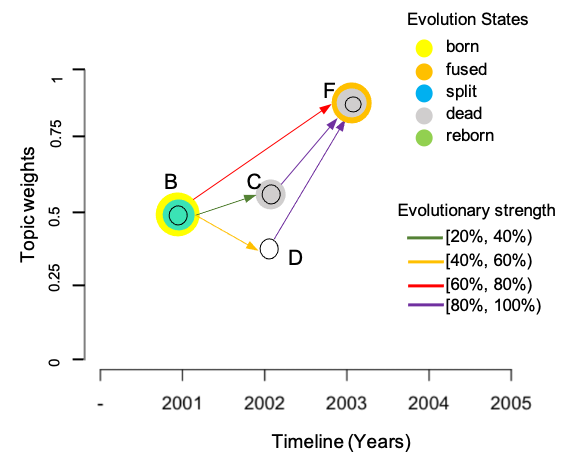}
            \subcaption{Before choosing the best parent of F}
        \end{minipage}
        \hfill
        \begin{minipage}[t]{.45\textwidth}
            \centering
            \includegraphics[trim=0cm 0cm 0cm 0cm, clip, width=200pt]{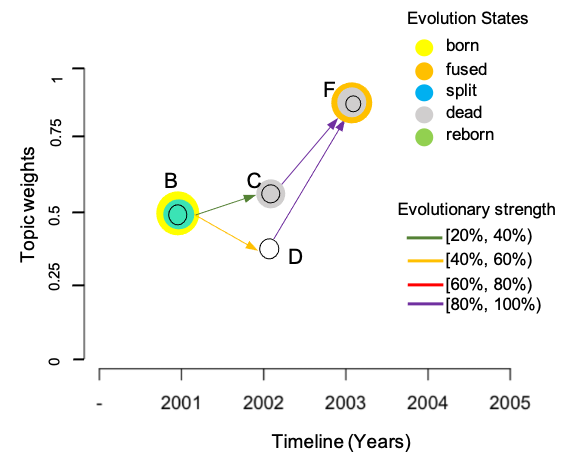}
            \subcaption{After choosing the best parent of F}
        \end{minipage}  
        \label{fig:1-2}
        \caption{An example of the trajectories.}\label{fig:multi_anc}
    \end{figure}

\end{enumerate}

\subsubsection{Topic trajectory visualiser}\label{sec:tet_visualiser}

Topic trajectory visualiser is implemented in the function \var{visualiseTET()} in \var{topicTracker.R}. The function has the following modules:

\begin{itemize}
    \item \textit{Create TET layout}: We create a TET using  \var{layout\_as\_tree()} (the Reingold-Tilford graph layout algorithm) in the R \var{igraph} package. We also define the positions of topics by their weights given by \var{weight} in $\mc{P}$ and time information given by $\var{year}$ in $\mc{P}$ within the layout.
    \item \textit{Determine evolution states of TET nodes}: We determine the evolution states of topics in the TET, and mark them using different colors (see below).
    \item \textit{Define R plotting function}: We define the \var{plot()} function: the properties of nodes and edges within the defined \var{layout\_as\_tree()} to visualise them in the TET.
    \item\textit{ Adjust overlapped labels of TET nodes}: The overlapped node labels in the TET are separated to avoid their overlaps using \var{thigmophobe.labels} in the package \var{plotrix}.
    \item \textit{Define x-axis and y-axis labels and visualize TET}: We finally define the ticks of x-axis and y-axis and draw their labels in the TET.
\end{itemize}

\ttv can identify the trajectories of the underlying topics, detecting their evolution states: (a) when and how a new topic is born? (b) how a topic can influence on the generation of newer topics? (c) what is a topic continually flourishing? (d) how a new topic can emerge implanted by what past topics? (e) what are the dead topics no longer observed? and (f) what are reborn topics distinguishable again in later time?
To address them, we measure the evolution states of underlying topics using the TESs given in $\mb{M}$:
\begin{itemize}
    \item \textit{born}: Topics {born} are the topics newly emerging. These are the topics emerging without any ancestors. 
    \item \textit{split}: Topics {split} are the topics split into more than two topics in the next generations. 
    \item \textit{fused}: Topics {fused} are the topics whose emergence has been made by more than two topics in the previous generations.
    \item \textit{reborn}: Topics {reborn} are the topics that re-emerge after  a user-specific time period, \minreborn. This state indicates that the topic {reborn} has attention after \minreborn, while during \minreborn it had been unnoticed. 
    % In a sense, \minreborn can be also seen as the `incubation period' of topics, where this period is the time elapsed between the moment of emergence of topics until their evolved topics appear.
    \item \textit{dead}: Topics {dead} are the topics that go into unobserved (being unpopular) during  \mindead. During \mindead, if a topic rarely influences the generation of topics in the next generation, we regard it as dead topics.
    \item \textit{flourishing}: Topics {flourishing} are the topics continually and actively being used or influencing on the generation of some topics in the next generations.
\end{itemize}

Note also that each topic can be marked by two evolutionary states: its emergence reason from the past generation, and its influence on the next generation. The former state is called `emerging-state' as it captures how it emerges. The latter is called `evolving-state' as it reveals its evolving state for topics in the next generations. 
% \begin{figure}[htb]
% \centering
% \includegraphics[trim=12cm 5cm 4cm 6cm, clip, width=500pt]{./image/tem_ex.pdf}
% \caption{\tem with example topics. By default, uncolored topics denote flourishing topics. The user-specified thresholds are set as: \minreborn = 2 years, \mindead = 1 years}\label{fig:tem_ex}
% \end{figure}
The evolution states are integrated with the TET, and this unified knowledge identifies the trajectories of the underlying topics. 
Finally, the underlying topics are positioned considering their weights, given by the \var{weight} field in $\mc{P}$, on the y-axis (see also Fig.~\ref{fig:ttv_overview}). This helps us to identify which topics more important than the others.

%% file: illustration.tex
\section{Illustrative examples}\label{sec:examples}
We present illustrative examples of \ttv which can also be reproduced in the provided Github URL.
Assume that the following temporal topic profile $\mc{P}$ in Table~\ref{tbl:P} and the TES matrix $\mb{M}$ in Table ~\ref{tbl:M} are given to \ttv. In $\mc{P}$, suppose that the \var{words} field contains the top-4 words for each topic.

\begin{table}[h!]
\centering
\scalebox{0.85}{
\setlength{\tabcolsep}{5pt}
\renewcommand{\arraystretch}{1}
\begin{tabular}{cccrcl} 
 \hline
id	&index&	label&	weight&	year&	words\\ 
\hline
t1&	0&	A&	0.75&	2001&	[`opinion', `computer', `lab', `user', `human']\\
t2&	1&	B&	0.5&	2001&	[`time', `lab', `opinion', `human', `computer']\\
t3&	2&	C&	0.55&	2002&	[`time', `application', `interface', `user', `computer']	\\
t4&	3&	D&	0.4&	2002&	[`lab', `user', `abc', `interface', `application']\\
t5&	4&	E&	0.1&	2002&	[`application', `interface', `abc', `opinion', `computer']\\
t6&	5&	F&	0.8&	2003&	[`interface', `computer', `application', `response', `system']\\
t7&	6&	G&	0.5&	2003&	[`lab', `abc', `survey', `opinion', `time']	\\
t8&	7&	H&	0.7&	2004&	[`machine', `lab', `system', `response', `human']\\
t9&	8&	I&	0.5&	2004&	[`opinion', `interface', `time', `application', `lab']\\
t10&	9&	J&	0.65&	2005&	[`opinion', `human', `user', `survey', `system']\\
t11&	10&	K&	0.45&	2005&	[`application', `interface', `system', `survey', `abc']\\
 \hline
\end{tabular}}
\caption{An example temporal topic profile $\mc{P}$. See also \var{softwarex\_example\_topic\_profile.csv} under the \var{data} directory of the Github URL.}
\label{tbl:P}
\end{table}

\begin{table}[h!]
\centering
\scalebox{0.9}{
\setlength{\tabcolsep}{4pt}
\renewcommand{\arraystretch}{1}
\begin{tabular}{c|rrrrrrrrrrr} 
& A & B & C &D&E&F&G&H&I&J&K\\ \hline
A&1&	0&	0.1&	0.1&	0.1&	0.9&	0.1&	0.7&	0.1&	0.1&	0.1\\
B&&	1&	0.3&	0.5&	0.1&	0.1&	0.1&	0.1&	0.1&	0.1&	0.1\\
C&&&		1&	0&	0&	0.1&	0.1&	0.1&	0.1&	0.1&	0.1\\
D&&&&			1&	0&	0.9&	0.1&	0.1&	0.1&	0.1&	0.1\\
E&&&&&				1&	0.2&	0.3&	0.1&	0.1&	0.1&	0.1\\
F&&&&&&					1&	0&	0.1&	0.1&	0.1&	0.1\\
G&&&&&&&						1&	0.1&	0.75&	0.1&	0.1\\
H&&&&&&&&							1&	0&	0.9	&0.9\\
I&&&&&&&&&								1&	0.1&	0.1\\
J&&&&&&&&&&									1&	0\\
K&&&&&&&&&&&										1\\
 \hline
\end{tabular}}
\caption{An example TES $\mb{M}$. The labels A-K are provided for the illustration only. In the actual input TES, only a  $N \times N$ matrix without the labels should be given, where $N$ is the number of the topics in $\mc{P}$. See also \var{softwarex\_example\_tes\_matrix.csv} under the \var{data} directory of the Github URL.}
\label{tbl:M}
\end{table}

As seen in Tables~\ref{tbl:P} - \ref{tbl:M}, there are 11 topics whose time slots are from 2001 to 2005. Given $\mc{P}$ and $\mb{M}$, \ttv creates the TES shown in Fig.~\ref{fig:tet_example}. We now discuss what insight we observe from this TET:

\begin{itemize}
    \item The label of each topic comes from the \var{label} string in Table~\ref{tbl:P}. If the user wants to display \var{id} and/or \var{words} as a label, the user can modify the code in \var{visualiseTES()}\footnote{We have commented which block needs to be modified in the code}.
    \item The topics at the same time slot are aligned together by the y-axis. The timeline is split by years 2001-2005. assuming that topics are captured and identified using a year. 
    \item The TESs are represented by the `Evolutionary strength' legend. For example, \var{A} influences the generation of both \var{F} and \var{H}, evidenced by \var{A} $\rightarrow$ \var{F} and \var{A} $\rightarrow$ \var{H}. The TES of \var{A} associated with \var{F} and \var{H} are assigned to the corresponding edges, respectively, using different colors. There is no meaning of the length of the edges.
    
    \item The evolutionary states of the topics are represented by different colors in the `Evolution states' legend. Each topic has two  evolution states: emerging-state and evolving-state.
    For example, the emerging-state of \var{A} is {born} as it newly emerges at 2001 without any ancestors. Its evolving-state is {split} as it influences the generation of both \var{F} and \var{H} as indicated by the directed edges. \var{H}'s emerging-state is {reborn}, with \minreborn = 2 years, meaning that \var{H} had been unnoticed during \minreborn from 2001 to 2003, but noticed again by being influenced by \var{A} after \minreborn at 2004. Its evolving-state is {split} as it influences on the generation of both \var{J} and \var{K}. 
    
    Uncolored topics represent flourishing. Both emerging-states and evolving-states of topics \var{D}, \var{G}, \var{I}, \var{J} and \var{K} are flourishing. \var{D} and \var{G} are influencing the generation of other topics in the next generations: \var{D} $\rightarrow$ \var{F}, \var{G} $\rightarrow$ \var{I}. \var{I}, \var{J} and \var{K} are under the incubation period which is less than \minreborn from the latest observed time slot, 2005.
    
    The emerging-state of \var{C} is flourishing as generated by the influence of \var{B}. Its evolving-state is \var{dead} as its influence had not been observed for the past 3 years from the latest time slot 2005, where the 3 years are greater than \mindead. The emerging-state of \var{F} is {fused} as generated by the co-influence of both \var{A} and \var{D}. Its evolving-state is {dead} with the same reason of that of \var{C}.
    
    \item The y-axis shows relative importance of topics. Referring to \var{B} $\rightarrow$ \var{D}, we see a declining trend of \var{B} (the weight decreases from 0.5 to 0.4) during 2001 - 2002, but that topic is getting more popular during 2002 - 2003 as indicated by \var{D} $\rightarrow$ \var{F} (the weight increases from 0.4 to 0.8). 
\end{itemize}

%% file: impact.tex
\section{Impact}\label{sec:impact}

Through a TET, we can observe the integrated, precious information about topic trajectories: evolutionary pathways of dynamic topics with evolution strengths indicated by the directed edges with different edge colors, evolution states of the topics indicated by different node colors, and the topic importance indicated by topic weight on the y-axis.
The goal of building a TET 
is to build the most likely genealogy of non-contemporary topics over time. This TET models topic trajectory information: the evolution pathways between non-contemporary topics based on their evolution strengths.

\ttv enlightens credible evolutionary relationships between non-contemporary topics. For example, in science and technology domains, \ttv could contribute to uncovering how technological or knowledge topics can change and influence the generation of newer topics through a series of evolution events over time. In e-commerce domains, \ttv could also help to track the evolution of market-competitive product-related topics in a product market over time.

\ttv is designed to generate topic trajectory identification and visualisation with few parameters. It can run with {two simple data formats} as explained in Section~\ref{sec:software}: a temporal topic profile, and a TES matrix. For any models that can provide these formats, their ability to identify topic trajectories can be easily analysed and visualised by \ttv. We believe that \ttv could help the user to allow a greater focus on methodological developments of their evolution strength matrix between time-stamped topics, rather than their implementation for topic trajectory identification. To the best of our knowledge, \ttv is the first generation of a framework that can be used in broader communities interested in identifying and visualising topic trajectory information. 

%% file: conclusion.tex
\section{Conclusions}\label{sec:conclusion}
In this paper, we presented a platform, \ttv, that can identify and visualise topic trajectory information. \ttv is equipped with the capability of addressing the issues still remained in the information retrieval community: how to identify trajectories of evolving topics over time? how to represent evolution states of the underlying topics at a particular time slot? and how to visualise topic trajectory information?
As the backbone of topic trajectory information, we presented that \ttv uses TET which aims to induce a most likely genealogy tree for evolving topics. We presented the formal definition of TET, its constituent elements, and detailed descriptions about how to construct a TET from two kinds of input data: a temporal topic profile  and a TES matrix showing the evolution strengths among the topics in the profile. Another key strength of \ttv is its ability to visualise three facets of useful trajectory information in a TET together: the evolutionary pathways of dynamic topics with their inter-evolution strengths, their evolution states at a particular time, and their relative importance. 
% We believe that \ttv could help users to focus on model development for topic trajectory identification rather than software development, and provide a platform for topic trajectory analysis that can promote ease-of-use. As future work, we plan to extend \ttv in the way that  enables a user to analyse TET properties and structure interactively that improves the capability of \ttv.
We believe that \ttv can provide a platform for topic trajectory analysis that can promote ease-of-use. 

% As future work, we plan to extend \ttv in the way that  enables a user to analyse TET properties and structure interactively that improves the capability of \ttv.

% As future work, we attract users in the information retrieval community to investigate and develop various algorithms for inferring scores for topic evolution strengths. By doing so, the community would get more benefit from enriched set of 

\section{Declaration of Competing Interest}
The authors declare that they have no known competing financial interests or personal relationships that could have appeared to influence the work reported in this paper.

%% file: ack.tex
\section*{Acknowledgements}

This work was supported by the LP170100416 project, funded by an Australian Research Council's Linkage grant. We thank Professor Beth Webster, Director of the Centre for Transformative Innovation at Swinburne University of Technology, for providing construct advice and support in developing \ttv. We  also thank both Dr. Don Klinkenberg and Dr. Thibaut Jombart who are the creators of the `phybreak' and `seqtrack' packages, respectively, for inspiring the foundation of \ttv. Special thanks to Dr. Don Klinkenberg for providing valuable suggestions for implementing \ttv.